# ON THE LOGIC OF CAUSAL MODELS *

Dan Geiger & Judea Pearl
Cognitive Systems Laboratory, Computer Science Department
University of California Los Angeles, CA 90024
Net address: geiger@cs.ucla.edu
Net address: judea@cs.ucla.edu

## ABSTRACT

This paper explores the role of Directed Acyclic Graphs (DAGs) as a representation of conditional independence relationships. We show that DAGs offer polynomially *sound* and *complete* inference mechanisms for inferring conditional independence relationships from a given *causal* set of such relationships. As a consequence, *d-separation,* a graphical criterion for identifying independencies in a DAG, is shown to uncover more valid independencies then any other criterion. In addition, we employ the *Armstrong* property of conditional independence to show that the dependence relationships displayed by a DAG are inherently *consistent,* i.e. for every DAG D there exists some probability distribution P that embodies all the conditional independencies displayed in D and none other.

## INTRODUCTION AND SUMMARY OF RESULTS

Networks employing Directed Acyclic Graphs (DAGs) have a long and rich tradition, starting with the geneticist Wright (1921). He developed a method called *path analysis* [Wright, 1934] which later on, became an established representation of causal models in economics [Wold, 1964], sociology [Blalock, 1971] and psychology [Duncan, 1975]. *Influence diagrams* represent another application of DAG representation [Howard and Matheson, 1981], [Shachter, 1988] and [Smith, 1987]. These were developed for decision analysis and contain both chance nodes and decision nodes (our definition of causal models excludes decision nodes). *Recursive models* is the name given to such networks by statisticians seeking meaningful and effective decompositions of contingency tables [Lauritzen, 1982], [Wermuth & Lauritzen, 1983], [Kiiveri et al, 1984]. *Bayesian Belief Networks* (or *Causal Networks*) is the name adopted for describing networks that perform evidential reasoning ([Pearl, 1986a, 1988]). This paper establishes a clear semantics for these networks that might explain their wide usage as models for forecasting, decision analysis and evidential reasoning.

DAGs are viewed as an economical scheme for representing conditional independence relationships. The nodes of a DAG represent variables in some domain and its topology is specified by a list of conditional independence judgements elicited from an expert in this domain. The specification list designates parents to each variable $v$ by asserting that $v$ is conditionally independent of all its predecessors, given its parents (in some total order of the variables). This input list implies many additional conditional independencies that can be read off the DAG. For example, the DAG asserts that, given its parents, $v$ is also conditionally independent of all its non-descendants [Howard and Matheson, 1981]. Additionally, if $S$ is a set of nodes containing $v$'s parents, $v$'s children and the parents of those children, then $v$ is in-

---

*This work was partially supported by the National Science Foundation Grant #IRI-8610155. "Graphoids: A Computer Representation for Dependencies and Relevance in Automated Reasoning (Computer Information Science)".

136

dependent of all other variables in the system, given those in $S$ [Pearl, 1986a]. These assertions are examples of *valid consequences* of the input list i.e., assertions that hold for every probability distribution that satisfies the conditional independencies specified by the input. If one ventures to perform topological transformations (e.g., arc reversal or node removal [Shachter, 1988]) on the DAG, caution must be exercised to ensure that each transformation does not introduce extraneous, invalid independencies, and/or that the number of valid independencies which become obscured by the transformation is kept at a minimum. Thus, in order to decide which transformations are admissible, one should have a simple graphical criterion for deciding which conditional independence statement is valid and which is not.

This paper deals with the following questions:

1.  What are the valid consequences of the input list ?

2.  What are the valid consequences of the input list that can be read off the DAG ?

3.  Are the two sets identical?

The answers obtained are as follows

1.  A statement is a valid consequence of the input set if and only if it can be derived from it using the axioms of semi-graphoids [Dawid, 1979; Pearl & Paz ,1985]. Letting $X$, $Y$, and $Z$ stand for three disjoint subsets of variables, and denoting by $I(X, Z, Y)$ the statement: " *the variables in $X$ are conditionally independent of those in $Y$, given those in $Z$* ", we may express these axioms as follows:

    Symmetry (1.a)
    $$I(X, Z, Y) \Rightarrow I(Y, Z, X)$$

    Decomposition (1.b)
    $$I(X, Z, Y \cup W) \Rightarrow I(X, Z, Y) \ \& \ I(X, Z, W)$$

    Weak Union (1.c)
    $$I(X, Z, Y \cup W) \Rightarrow I(X, Z \cup W, Y)$$

    Contraction (1.d)
    $$I(X, Z \cup Y, W) \ \& \ I(X, Z, Y) \Rightarrow I(X, Z, Y \cup W)$$

2.  Every statement that can be read off the DAG using the d-separation criterion is a valid consequence of the input list [Verma, 1986].

    The $d$-separation condition is defined as follows [Pearl, 1985]: For any three disjoint subsets $X, Y, Z$ of nodes in a DAG D, $Z$ is said to $d$-*separate* $X$ from $Y$, denoted $I(X, Z, Y)_D$, if there is no path from a node in $X$ to a node in $Y$ along which: 1. every node that delivers an arrow is outside $Z$, and 2. every node with converging arrows either is in $Z$ or has a descendant in $Z$ (the definition is elaborated in the next section).

3.  The two sets are identical, namely, a statement is valid IF AND ONLY IF it is graphically-validated under d-separation in the DAG.

137

The first result establishes the *decidability* of verifying whether an arbitrary statement is a valid consequence of the input set. The second result renders the d-separation criterion a *polynomially sound* inference rule, i.e., it runs in polynomial time and certifies only valid statements. The third renders the d-separation criterion a *polynomially complete* inference rule, i.e., the DAG constitutes a sound and complete inference mechanism that identifies, in polynomial time, each and every valid consequence in the system.

The results above are true only for *causal* input sets i.e., those that recursively specify the relation of each variable to its predecessors in some (chronological) order. The general problem of verifying whether a given conditional independence statement logically follows from an arbitrary set of such statements, may be undecidable. Its decidability would be resolved upon finding a complete set of axioms for conditional independence i.e., axioms that are powerful enough to derive all valid consequences of an arbitrary input list. The completeness problem is treated in [Geiger & Pearl, 1988] and completeness results for specialized subsets of probabilistic dependencies have been obtained. All axioms encountered so far are derivable from Dawid's axioms, which suggests that they are indeed complete, as conjectured in [Pearl & Paz, 1985]. Result-1 can be viewed as yet another completeness result for the special case in which the input statements form a causal set. This means that applying axioms (1.a) through (1.d) on a causal input list is guaranteed to generate all valid consequences and none other. Interestingly, result-2 above holds for any statements that obey Dawid's axioms, not necessarily probabilistic conditional independencies. Thus, DAGs can serve as polynomially sound inference mechanisms for a variety of dependence relationships, e.g., partial correlations and qualitative database dependencies. In fact, the results of this paper prove that $d$-separation is complete for partial correlation as well as for conditional independence statements, whereas completeness for qualitative database dependencies has not been examined.

## SOUNDNESS AND COMPLETENESS

The definition of d-separation is best motivated by regarding DAGs as a representation of causal relationships. Designating a node for every variable and assigning a link between every cause to each of its direct consequences defines a graphical representation of a causal hierarchy. For example, the propositions "It is raining" ($\alpha$), "the pavement is wet" ($\beta$) and "John slipped on the pavement" ($\gamma$) are well represented by a three node chain, from $\alpha$ through $\beta$ to $\gamma$; it indicates that either rain or wet pavement could cause slipping, yet wet pavement is designated as the *direct cause;* rain could cause someone to slip if it wets the pavement, but not if the pavement is covered. Moreover, knowing the condition of the pavement renders "slipping" and "raining" independent, and this is represented graphically by a $d$-separation condition, $I(\alpha, \gamma, \beta)_D$, showing node $\alpha$ and $\beta$ separated from each other by node $\gamma$. Assume that "broken pipe" ($\delta$) is considered another direct cause for wet pavement, as in figure 1. An induced dependency exists between the two events that may cause the pavement to get wet: "rain" and "broken pipe". Although they appear connected in Figure 1, these propositions are marginally independent and become dependent once we learn that the pavement is wet or that someone broke his leg. An increase in our belief in either cause would decrease our belief in the other as it would "explain away" the observation. The following definition of $d$-separation permits us to graphically identify such induced dependencies from the DAG ($d$ connoted "directional").

**Definition:** If $X$, $Y$, and $Z$ are three disjoint subsets of nodes in a DAG D, then $Z$ is said to *d-separate* $X$

from $Y$, denoted $I(X,Z,Y)_D$, iff there is no path* from a node in $X$ to a node in $Y$ along which every node that delivers an arrow is outside $Z$ and every node with converging arrows either is or has a descendant in $Z$. A path satisfying the conditions above is said to be *active*, otherwise it is said to be *blocked* (by $Z$). Whenever a *statement* $I(X,Z,Y)_D$ holds in a DAG $D$, the predicate $I(X,Z,Y)$ is said to be *graphically-verified* (or an *independency*), otherwise it is *graphically-unverified* by $D$ (or a *dependency*).

In figure 2, for example, $X=\{2\}$ and $Y=\{3\}$ are $d$-separated by $Z=\{1\}$; the path $2 \leftarrow 1 \rightarrow 3$ is blocked by $1 \in Z$ while the path $2 \rightarrow 4 \leftarrow 3$ is blocked because 4 and all its descendents are outside $Z$. Thus $I(2,1,3)$ is graphically-verified by $D$. However, $X$ and $Y$ are not $d$-separated by $Z'=\{1,5\}$ because the path $2 \rightarrow 4 \leftarrow 3$ is rendered active. Consequently, $I(2,\{1,5\},3)$ is graphically-unverified by $D$; by virtue of 5, a descendent of 4, being in $Z$. Learning the value of the consequence 5, renders its causes 2 and 3 dependent, like opening a pathway along the converging arrows at 4.

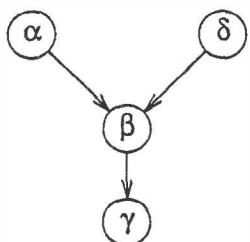

Figure 1

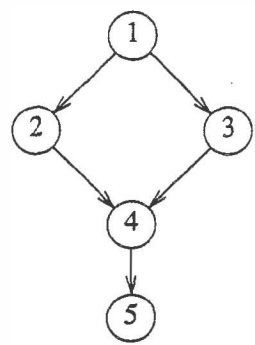

Figure 2

**Definition:** If $X$, $Y$, and $Z$ are three disjoint subsets of variables of a distribution $P$, then $X$ and $Y$ are said to be conditionally independent given $Z$, denoted $I(X,Z,Y)_P$ iff $P(X,Y \mid Z) = P(X \mid Z) \cdot P(Y \mid Z)$ for all possible values of $X$, $Y$ and $Z$ for which $P(Z) > 0$. $I(X,Z,Y)_P$ is called a *(conditional independence) statement*. A conditional independence statement $\sigma$ *logically follows* from a set $\Sigma$ of such statements if $\sigma$ holds in every distribution that obeys $\Sigma$. In such case we also say that $\sigma$ is a *valid consequence* of $\Sigma$.

Ideally, to employ a DAG $D$ as a graphical representation for dependencies of some distribution $P$ we would like to require that for every three disjoint sets of variables in $P$ (and nodes in $D$) the following equivalence would hold

$$I(X,Z,Y)_D \quad \textit{iff} \quad I(X,Z,Y)_P \tag{2}$$

This would provide a clear graphical representation of all variables that are conditionally independent. When equation (2) holds, $D$ is said to be a *perfect map* of $P$. Unfortunately, this requirement is often too strong because there are many distributions that have no perfect map in DAGs. The spectrum of probabilistic dependencies is in fact so rich that it cannot be cast into any representation scheme that uses polynomial amount of storage ([Verma, 1987]). Geiger [1987] provides a graphical representation based on a collection of graphs (Multi-DAGs) that is powerful enough to perfectly represent an arbitrary distribution, however, as shown by Verma, it requires, on the average, an exponential number of DAGs. Being unable to provide perfect maps at a reasonable cost, we compromise the requirement that the graphs represent each and every dependency of $P$, and allow some independencies to escape representation.

---

* By *path* we mean a sequence of edges in the underlying undirected graph, i.e ignoring the directionality of the links.

**Definition:** A DAG $D$ is said to be an *I-map* of $P$ if for every three disjoint subsets $X$, $Y$ and $Z$ of variables the following holds:

$$I(X,Z,Y)_D \Rightarrow I(X,Z,Y)_P$$

The natural requirement for these I-maps is that the number of undisplayed independencies be minimized.

The task of finding a DAG which is a minimal I-map of a given distribution $P$ was solved in [Verma 1986; Pearl & Verma, 1987]. The algorithm consists of the following steps: assign a total ordering $d$ to the variables of $P$. For each variable $i$ of $P$, identify a minimal set of predecessors $S_i$ that renders $i$ independent of all its other predecessors (in the ordering of the first step). Assign a direct link from every variable in $S_i$ to $i$. The resulting DAG is an I-map of $P$, and is minimal in the sense that no edge can be deleted without destroying its I-mapness. The input list $L$ for this construction consists of $n$ conditional independence statements, one for each variable, all of the form $I(i, S_i, U_{(i)}-S_i)$ where $U_{(i)}$ is the set of predecessors of $i$ and $S_i$ is a subset of $U_{(i)}$ that renders $i$ conditionally independent of all its other predecessors. This set of conditional independence statements is called a *causal* input list and is said to *define* the DAG D. The term "causal" input list stems from the following analogy: Suppose we order the variables chronologically, such that a cause always precedes its effect. Then, from all potential causes of an effect $i$, a causal input list selects a minimal subset that is sufficient to explain $i$, thus rendering all other preceding events superfluous. This selected subset of variables are considered the *direct causes* of $i$ and therefore each is connected to it by a direct link.

Clearly, the constructed DAG represents more independencies than those listed in the input, namely, all those that are graphically verified by the d-separation criterion. The analysis of [Verma, 1986] guarantees that all graphically-verified statements are indeed valid in $P$ i.e., the DAG is an I-map of $P$. However, this paper shows that the constructed DAG has an additional property; it graphically-verifies **every** conditional independence statement that logically follows from $L$ (i.e. holds in every distribution that obeys $L$). Hence, we cannot hope to improve the $d$–*separation* criterion to display more independencies, because all valid consequences of $L$ (which defines $D$) are already captured by $d$-separation.

The three theorems below formalize the above discussion.

**Theorem 1 (soundness) [Verma, 1986]:** Let $D$ be a DAG defined by a causal input list $L$. Then, every graphically-verified statement is a valid consequence of $L$.

**Theorem 2 (closure) [Verma, 1986]:** Let $D$ be a DAG defined by a causal input list $L$. Then, the set of graphically-verified statements is exactly the closure of $L$ under axioms (1.a) through (1.d).

**Theorem 3 (completeness):** Let $D$ be a DAG defined by a causal input list $L$. Then, every valid consequence of $L$ is graphically-verified by $D$ (equivalently, every graphically-unverified statement in $D$ is not a valid consequence of $L$).

140

Theorem 1 guarantees that the DAG displays only valid statements. Theorem 2 guarantees that the DAG displays all statements that are derivable from $L$ via axioms (1). The third theorem, which is the main contribution of this paper, assures that the DAG displays all statements that logically follow from $L$ i.e., the axioms in (1) are complete, capable of deriving all valid consequences of a causal input list. Moreover, since a statement in a DAG can be verified in polynomial time, theorem 3 provides a complete polynomial inference mechanism for deriving all independency statements that are implied by a causal input set.

Theorem 3 is proven in the appendix by actually constructing a distribution $P_\sigma$ that satisfies all conditional independencies in $L$ and violates any statement $\sigma$ graphically-unverified in $D$. This distribution precludes $\sigma$ from being a valid consequence of $L$ and therefore, since the construction can be repeated for every graphically-unverified statement, none of these statements is a valid consequence of $L$.

The first two theorems are more general than the third in the sense that they hold for every dependence relationship that obeys axioms (1.a) through (1.d), not necessarily those based on probabilistic conditional independence (proofs can be found in [Verma, 1986]). Among these dependence relationships are partial correlations ([Pearl & Paz, 1985]) and qualitative dependencies ([Fagin, 1982], [Shafer at al, 1987]) which can readily be shown to obey axioms (1). Thus, for example, the transformation of arc-reversal and node removal ([Howard & Matheson, 1981]) can be shown valid by purely graphical consideration, simply showing that every statement verified in the transformed graph is also graphically-verified in the original graph.

The proof of theorem 3 assumes that $L$ contains only statements of the form $I(i, S_i, U_{(i)}-S_i)$. Occasionly, however, we are in possession of stronger forms of independence relationships, in which case additional statements should be read of the DAG. A common example is the case of a variable that is functionally dependent on its corresponding parents in the DAG ( *deterministic variable,* [Shachter, 1988]). The existence of each such variable $i$ could be encoded in $L$ by a statement of *global* independence $I(i, S_i, U-S_i-i)$ asserting that conditioned on $S_i$, $i$ is independent of all other variables, not merely of its predecessors. The independencies that are implied by the modified input list can be read from the DAG using an enhanced version of $d$-separation, named *ID-separation.*

**Definition:** If $X$, $Y$, and $Z$ are three disjoint subsets of nodes in a DAG D, then $Z$ is said to *ID-separate* $X$ from $Y$, iff there is no path from a node in $X$ to a node in $Y$ along which 1. every node which delivers an arrow is outside $Z$, 2. every node with converging arrows either is or has a descendant in $Z$ and 3. no node is functionally determined by $Z$.

The new criterion certifies all independencies that are revealed by $d$-separation plus additional ones due to the enhancement of the input list. It has been shown that theorem 1 through 3 hold for *ID*-separation whenever $L$ contains global independence statements [Geiger & Verma, 1988].

These graphical criteria provide easy means of recognizing conditional independence in influence diagrams as well as identifying the set of parameters needed for any given computation. Shachter [1985, 1988] has devised an algorithm for finding a set of nodes $M$ guaranteed to contain sufficient information for computing $P(x| y)$, for two arbitrary sets of variables $x$ and $y$. The outcome of Shachter's algorithm can now be stated declaratively; $M$ contains every ancestor of $x \cup y$ that is not ID-separated from $x$ given $y$ and none other. The completeness of ID-separation implies that $M$ is minimal; no node in $M$ can be excluded on purely topological grounds (i.e., without considering the numerical values of the probabilities involved).

141

We conclude by showing how these theorems can be employed as an inference mechanism. Assume an expert has identified the following conditional independencies between variables denoted 1 through 5:

$$L = \{ I(2, 1, \varnothing), I(3, 1, 2), I(4, 23, 1), I(5, 4, 123) \}$$

(the first statement in $L$ is trivial). We address two questions. First, what is the set of all valid consequences of $L$? Second, in particular, is $I(3, 124, 5)$ a valid consequence of $L$? For general input lists the answer for such questions may be undecidable but, since $L$ is a causal list, it defines a DAG that graphically verifies each and every valid consequences of $L$. The DAG $D$ is the one shown in figure 2, which constitutes a dense representation of all valid consequences of $L$. To answer the second question, we simply observe that $I(3, 124, 5)$ is graphically-verified in $D$. A graph-based algorithm for another subclass of statements, called *fixed context* statements, is given in [Geiger & Pearl, 1988]. In that paper, results analogous to theorem 1 through 3 are proven for Markov-fields; a representation scheme based on undirected graphs ([Isham, 1981], [Lauritzen, 1982]).

## EXTENSIONS AND ELABORATIONS

Theorem 3 can be restated to assert that for every DAG D and any dependency $\sigma$ there exist a probability distribution $P_\sigma$ that satisfies D's input set $L$ and the dependency $\sigma$. By theorem 2, $P_\sigma$ must satisfy all graphically-verified statements as well because they are all derivable from $L$ by Dawid's axioms. Thus, theorems 2 and 3 guarantee the existence of a distribution $P_\sigma$ that satisfies all graphically verified statements and a single arbitrary-chosen dependency. The question answered in this section is the existence of a distribution $P$ that satisfies all independencies of $D$ and all its dependencies (not merely a single dependency). We show that such a distribution exists, which legitimizes the use of DAGs as a representation scheme for probabilistic dependencies; a model builder who uses the language of DAGs to express dependencies is guarded from inconsistencies.

The construction of $P$ is based on the Armstrong property of conditional independence.

**Definition:** Conditional independence is an *Armstrong relation* in a class of distributions **P** if there exists an operation $\otimes$ that maps finite sequences of distributions in **P** into a distribution of **P**, such that if $\sigma$ is a conditional independence statement and if $P_i$ $i=1..n$ are distributions in **P**, then $\sigma$ holds for $\otimes \{P_i \mid i=1..n\}$ iff $\sigma$ holds for each $P_i$.

The notion of Armstrong relation is borrowed from database theory ([Fagin 1982]). We concentrate on two families of distributions **P**: All distributions, denoted $PD$ and strictly positive distributions, denoted $PD^+$. Conditional independence can be shown to be an Armstrong relation in both families. The construction of the operation $\otimes$ is given below, however the proof is omitted and can be found in ([Geiger & Pearl, 1988]).

**Theorem 4 ([Geiger & Pearl, 1988]):** Conditional independence is an Armstrong relation in $PD$ and in $PD^+$.

We shall construct the operation $\otimes$ for conditional independence using a binary operation $\otimes'$ such that if $P = P_1 \otimes' P_2$ then for every conditional independency statement $\sigma$ we get

$$\otimes' P_i \text{ obeys } \sigma \quad \text{iff} \quad P_1 \text{ obeys } \sigma \text{ and } P_2 \text{ obeys } \sigma. \tag{5}$$

The operation $\otimes$ is recursively defined in terms of $\otimes'$ as follows:

$$\otimes \{ P_i \mid i=1..n \} = ((P_1 \otimes' P_2) \otimes' P_3) \otimes' \cdots P_n ).$$

Clearly, if $\otimes'$ satisfies equation (5), then $\otimes$ satisfies the the requirement of an Armstrong relation, i.e.

$$P \text{ obeys } \sigma \quad \text{iff} \quad \forall_i P_i \text{ obeys } \sigma.$$

Therefore, it suffices to show that $\otimes'$ satisfies (5).

Let $P_1$ and $P_2$ be two distributions sharing the variables $x_1, \cdots, x_n$. Let $A_1, \cdots, A_n$ be the domains of $x_1, \cdots, x_n$ in $P_1$ and let an instantiation of these variables be $\alpha_1, \cdots, \alpha_n$. Similarly, let $B_1, \cdots, B_n$ be the domains of $x_1, \cdots, x_n$ in $P_2$ and $\beta_1, \cdots, \beta_n$ an instantiation of these variables. Let the domain of $P = P_1 \otimes' P_2$ be the product domain $A_1 B_1, \cdots, A_n B_n$ and denote an instantiation of the variables of $P$ by $\alpha_1 \beta_1, \cdots, \alpha_n \beta_n$. Define $P_1 \otimes' P_2$ by the following equation:

$$P(\alpha_1 \beta_1, \alpha_2 \beta_2, \cdots, \alpha_n \beta_n) = P_1(\alpha_1, \alpha_2, \cdots, \alpha_n) \cdot P_2(\beta_1, \beta_2, \cdots, \beta_n).$$

The proof that $P$ satisfies the condition of theorem 4 uses only the definition of conditional independence and can be found in [Geiger & Pearl 1988]. The adequacy of this construction for $PD^+$ is due to the fact that $\otimes$ produces a strictly positive distribution whenever the input distributions are strictly positive.

**Theorem 5:** For every DAG D there exists a distribution $P$ such that for every three disjoint sets of variables $X$, $Y$ and $Z$, the following holds;

$$I(X,Z,Y)_D \quad \text{iff} \quad I(X,Z,Y)_P$$

**Proof:** Let $P = \otimes \{ P_\sigma \mid \sigma \text{ is a dependency in a DAG D} \}$ where $P_\sigma$ is a distribution obeying all independencies of $D$ and a dependency $\sigma$. By theorem 3, a distribution $P_\sigma$ always exists. $P$ satisfies the requirement of theorem 5 because it obeys only statements that hold in every $P_\sigma$ and these are exactly the ones verified by $D$. $\square$

The construction presented in the proof of theorem 5 leads to a rather complex distribution, where the domain of each variable is unrestricted. It still does not guarantee that a set of dependencies and independencies represented by DAGs is realizable in a more limited class of distributions such as normal or those defined on binary variables. We conjecture that these two classes of distributions are sufficiently rich to permit the consistency of DAG representation.

## ACKNOWLEDGMENT

We thank Ron Fagin for pointing out the usefulness of the notion of Armstrong relation and to Ross Shachter for his insightful comments. Thomas Verma and Azaria Paz provided many useful discussions on the properties of dependency models.

# APPENDIX

**Theorem 3 (completeness):** Let $D$ be a DAG defined by a causal input list $L$. Then, every valid consequence of $L$ is graphically-verified by $D$.

**Proof:** Let $\sigma = I(X, Z, Y)$ be an arbitrary graphically-unverified statement in $D$. We construct a distribution $P_\sigma$ that satisfies all conditional independencies in the input list $L$ and violates $\sigma$. This distribution precludes $\sigma$ from being a valid consequence of $L$ and therefore, every valid consequence of $L$ must be graphically-verified in $D$.

>From the definition of $d$-separation, there must exist an active path between an element $\alpha$ in $X$ and an element $\beta$ in $Y$ that is not d-separated by $Z$. Ensuring that $P_\sigma$ violates the conditional independency $I(\alpha, Z, \beta)$, denoted $\sigma'$, guarantees that $\sigma$ is also violated, because any distribution that renders $X$ and $Y$ conditionally independent must render each of their individual variables independent as well (axiom (1.b)).

$P_\sigma$ is defined in terms of a simplified DAG $D_\sigma$. This DAG is constructed by removing as many links as possible from $D$ such that $\sigma$ remains unverified in $D_\sigma$. This process clearly preserves all previously verified statements but caution is exercised not to remove links that would render $\sigma$ graphically-verified in $D_\sigma$. We will conclude the proof by constructing a distribution $P_\sigma$ which satisfies all graphically-verified statements of $D_\sigma$ (hence also those of $D$) and violates $\sigma'$.

Let $q$ be an active path (by $Z$) between $\alpha$ and $\beta$ with the minimum number of head-to-head nodes (i.e. nodes with converging arrows) denoted, left to right, $h_1, h_2, ..., h_k$. Let $z_i$ be the closest (wrt path length) descendent of $h_i$ in $Z$ and let $p_i$ be the directed path from $h_i$ to $z_i$ (if $h_i \in Z$ then $z_i = h_i$). We define $D_\sigma$ to be a subgraph of $D$ containing only the links that form the paths $p_i$'s and the path $q$. We make two claims about the topology of the resulting DAG. First, the paths $p_i$ are all distinct. Second, for any $i$, $h_i$ is the only node shared by $p_i$ and $q$. The resulting DAG is depicted in figure 3 (note that some nodes, including nodes of $Z$, might become isolated in $D_\sigma$).

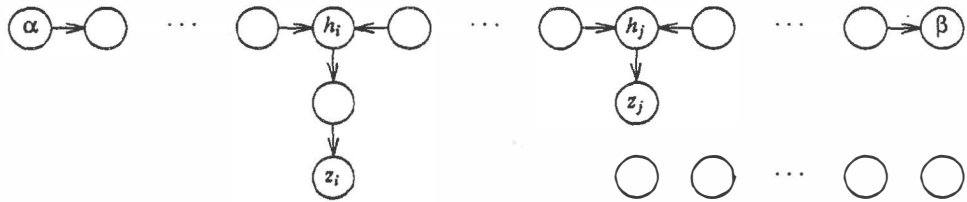

*Figure 3*

Proof of claim 1: Assume, by contradiction, that there are two paths $p_i$ and $p_j$ ($i < j$) with a common node $\gamma$. Under this assumption, we find an active path between $\alpha$ and $\beta$ that has less head-to-head nodes then $q$, contradicting the minimality of the latter. If $\gamma$ is neither $h_i$ nor $h_j$ then the path $(\alpha, h_i, \gamma, h_j, \beta)$ is an active path (by $Z$); Each of its head-to-head nodes is or has a descendent in $Z$ because it is either $\gamma$ or a head-to-head node of $q$. Every other node lies either on the active path $q$ and therefore is outside $Z$ or lies on $p_i$ ($p_j$) in which case, since it has a descendent $\gamma$, it must also be outside $Z$. The resulting path contradicts the minimality of $q$ since both $h_i$ and $h_j$ are no longer head-to-head nodes while $\gamma$ is the only newly introduced head-to-head node. If $\gamma = h_j$ then, using similar arguments, the path $(\alpha, h_i, \gamma, \beta)$ (see figure 4), which has less head-to-head nodes then $q$, can readily be shown active (the case $\gamma = h_i$ is similar).

144

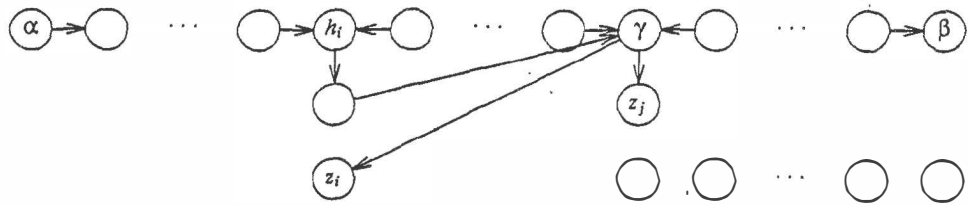

*Figure 4*

Proof of claim 2: Assume $p_i$ and $q$ have in common a node $\gamma$ other then $h_i$ and assume w.l.o.g that it lies between $h_i$ and $\beta$. This node is not a head-to-head node on $q$ because $p_i$ is distinct from all other $p_j$'s. The node $\gamma$ cannot belong to $Z$ because otherwise $q$ would not have been active. Thus, the path $(\alpha, h_i, \gamma, \beta)$ must be an active path which contradicts the minimality of $q$ (figure 5).

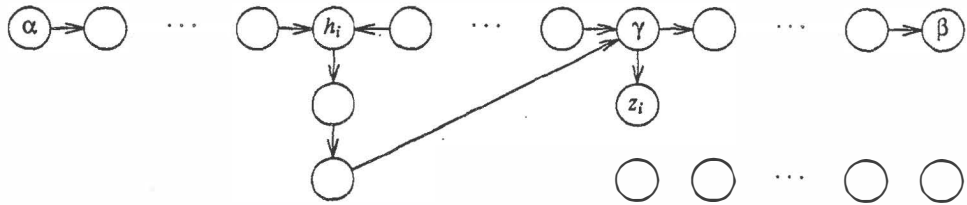

*Figure 5*

In the following discussion we call a path containing no head-to-head node a *regular path*. Let $P_\sigma$ be a normal distribution with the following covariance matrix $\Gamma$

$$\Gamma = (\rho_{ij}) \quad \rho_{ij} = \begin{cases} 0 & \text{There exists no regular path from node } i \text{ to node } j \\ \rho^l & \text{There exists a regular path of length } l \text{ between node } i \text{ and node } j \end{cases} \quad (3)$$

Since $D_\sigma$ is singly connected there exist at most one path between any two nodes. Any value of $\rho$ satisfying $0 < n \cdot \rho^2 < 1$ would render $\Gamma$ positive definite and therefore a valid covariance matrix. We claim that this distribution satisfies all independencies of $D_\sigma$ and violates $I(\alpha, Z, \beta)$. To evaluate $I(\alpha, Z, \beta)$ we first form the projection of $P_\sigma$ on the variables $\alpha$, $\beta$ and $Z$. Since $P_\sigma$ is normal, this projection is also normal and its covariance matrix is a submatrix $\Gamma'$ of $\Gamma$ that corresponds to the variables $\alpha$, $\beta$ and $Z$. The statement $I(\alpha, Z, \beta)$ holds in $P_\sigma$ iff $det(\Gamma'_{\alpha\beta}) = 0$ where $\Gamma'_{\alpha\beta}$ is a submatrix of $\Gamma'$ obtained by removing the $\alpha$-th line and the $\beta$-th column ([Miller, 1964]). Both $\Gamma'$ and $\Gamma'_{\alpha\beta}$ are given below. The matrix $\Gamma'$ is a tri-diagonal matrix whose off main-diagonal elements are integer powers of $\rho$ or zeros. The columns and lines of $\Gamma'$ correspond to the following order of variables: $\alpha$, $z_1,...,z_k$, $\beta$ and then all other variables of $Z$ (see fig 3), thus, for example, the term $\rho^{i_1}$ located at (1,2) in $\Gamma'$ is the correlation factor between $\alpha$ and $z_1$ because these variables are the first two in the above order. The integer $i_1$ is the length of the path between $\alpha$ and $z_1$, $i_2$ is the length of the path between $z_1$ and $z_2$ and so on. The construction of $\Gamma'_{\alpha\beta}$ is based on the observation that the location $(\alpha, \beta)$ in $\Gamma'$ is $(1, k+2)$ where $k$ is the number of head-to-head nodes of $q$.

$$\Gamma' = \begin{bmatrix} 1 & \rho^{i_1} & 0 & 0 & 0 \\ \rho^{i_1} & 1 & \rho^{i_2} & 0 & 0 \\ 0 & \rho^{i_2} & 1 & \rho^{i_3} & 0 \\ 0 & 0 & \rho^{i_3} & 1 & 0 \\ 0 & 0 & 0 & 0 & 1 \end{bmatrix} \quad \Gamma'_{\alpha\beta} = \begin{bmatrix} \rho^{i_1} & 1 & \rho^{i_2} & 0 \\ 0 & \rho^{i_2} & 1 & 0 \\ 0 & 0 & \rho^{i_3} & 0 \\ 0 & 0 & 0 & 1 \end{bmatrix}$$

(These matrices are given for the case of two head-to-head nodes and a single additional isolated node of $Z$, their general form is obvious). Clearly, $det(\Gamma'_{\alpha\beta}) = \rho^k$ ($k \geq 0$) and therefore chosing $\rho \neq 0$ guarantees that $I(\alpha, Z, \beta)$ does not hold in $P_\sigma$.

It remains to show that every graphically-verified statement in $D_\sigma$ is satisfied by $P_\sigma$. We assign a total order $d$ on the nodes of $D_\sigma$ consistent with the partial order determined by $D_\sigma$. We show that the $n$ statements that form the causal input list that defines $D_\sigma$ are satisfied in $P_\sigma$. Theorem 1 ensures that all other graphically verified statements are valid consequences of this input list and therefore would all be satisfied in $P_\sigma$. In what follows we use the tag of a node as its name. Let $I(i, S_i, U_{(i)}-S_i)$ be a statement of $L$ where $S_i$ are the parents of $i$ and $U_{(i)}$ are all the variables preceding $i$ in the order $d$. By the topology of $D_\sigma$, $S_i$ contains no more then two nodes.

**Assume $S_i$ is empty.** This implies that $i$ is not connected via a regular path to either of its predecessors. Hence, by the construction of $\Gamma$, $\rho_{ij} = 0$ for every $j \in U_{(i)}$, and therefore the statements $I(i, \emptyset, j)$ hold in $P_\sigma$. However, in normal distributions, the correlation between single variables determines the dependency between the sets containing these variables because the following axiom holds.

(Composition-Decomposition)

$$I(X, Z, Y \cup W) \iff I(X, Z, Y) \,\&\, I(X, Z, W) \qquad (4)$$

Accordingly, $I(i, \emptyset, U_{(i)})$ holds in $P_\sigma$ and this statement is exactly equal to $I(i, S_i, U_{(i)}-S_i)$ since $S_i$ is empty.

**Assume $S_i$ consists of a single node $h$.** In light of axiom (4) it is enough to show that for every $j \in U_{(i)}-h$ the statement $I(i, h, j)$ holds in $P_\sigma$. If there exists a path from $j$ to $i$, it must pass through $h$. Therefore, by definition (3) of $\Gamma$, since $h$ is the only parent of $i$, the equality $\rho_{ij} = \rho_{ih} \rho_{hj}$ must be satisfied. This equality is a necessary and sufficient condition under which $I(i, h, j)$ holds in any normal distribution in which $\rho_{ij}$ are the correlation factors (in particular, $P_\sigma$).

**Assume $S_i = \{g, h\}$** (see figure 6). Again, it is enough to show that $I(i, \{g, h\}, j)$ holds in $P_\sigma$ for every $j \in U_{(i)}-S_i$. Construct the covariance matrix for the variables $g, h, i$ and $j$ (the columns of the matrix correspond to this order). By equation (3), $\rho_{gi} = \rho$, $\rho_{hi} = \rho$ and $\rho_{gh} = 0$. The resulting matrix is given below,

$$\Gamma' = \begin{bmatrix} 1 & 0 & \rho & \rho_{gj} \\ 0 & 1 & \rho & \rho_{hj} \\ \rho & \rho & 1 & \rho_{ij} \\ \rho_{gj} & \rho_{hj} & \rho_{ij} & 1 \end{bmatrix}.$$

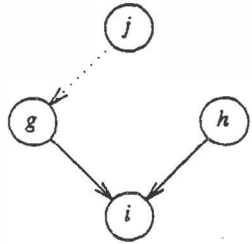

*Figure 6*

The statement $I(i, \{g, h\}, j)$ holds in this distribution iff the submatrix $\Gamma_{ij}'$ is singular, i.e.

146

$$\det \begin{bmatrix} 1 & 0 & \rho \\ 0 & 1 & \rho \\ \rho_{gj} & \rho_{hj} & \rho_{ij} \end{bmatrix} = 0$$

([Miller, 1964]) or equivalently, $\rho_{ij} = (\rho_{hj} + \rho_{gj}) \cdot \rho$.

The latter equality, however, holds for all possible selections of a node $j$; If $j$ is not connected to $i$ via a regular path, i.e. $\rho_{ij} = 0$, then it is not connected through a regular path to either of $i$'s parents and therefore both $\rho_{gj}$ and $\rho_{hj}$ are zero. If $j$ is connected through a regular path of length $l$ to $g$ (similarly when connected to $h$) then it is connected to its son $i$ with a path of length $l+1$ and is not connected to $i$'s other parent, in which case $\rho_{ij} = \rho^{l+1}$, $\rho_{gj} = \rho^{l}$, $\rho_{hj} = 0$ and therefore the above equality holds. □.